\documentclass[11pt]{article}
\usepackage{eamt20}
\usepackage{times}
\usepackage{url}
\usepackage{latexsym}
\usepackage[small,bf]{caption} 
\usepackage[english]{babel}
\usepackage[utf8]{inputenc}

\usepackage{latexsym}

\usepackage{amsmath, amsthm, amssymb}
\usepackage{makecell}

\usepackage{tikz, tikz-qtree}
\usetikzlibrary{arrows}
\usetikzlibrary{fit} 

\usepackage{algorithm}
\usepackage{algpseudocode}

\usepackage{forest}

\usepackage{multirow}
\usepackage{graphicx}

\title{Learning synchronous context-free grammars with multiple specialised non-terminals for hierarchical phrase-based translation$^{*}$}

\author{
{\bf Felipe Sánchez-Martínez,
Juan Antonio Pérez-Ortiz,
\bf Rafael C. Carrasco}\\[1ex]
Dep. de Llenguatges i Sistemes Informàtics,
Universitat d'Alacant, Spain \\
{\tt \{fsanchez,japerez,carrasco\}@dlsi.ua.es}\\[1ex]
}

\date{}

\begin{document}
\maketitle

\begin{abstract}
  Translation models based on hierarchical phrase-based statistical machine translation (HSMT) have shown better performances than the non-hierarchical phrase-based counterparts for some language pairs. The standard approach to HSMT learns and apply a synchronous context-free grammar with a single non-terminal. The hypothesis behind the grammar refinement algorithm presented in this work is that this single non-terminal is overloaded, and insufficiently discriminative, and therefore, an adequate split of it into more specialised symbols could lead to improved models. This paper presents a method to learn synchronous context-free grammars with a huge number of initial non-terminals, which are then grouped via a clustering algorithm. Our experiments show that the resulting smaller set of non-terminals correctly capture the contextual information that makes it possible to statistically significantly improve the BLEU score of the standard HSMT approach.
\end{abstract}

\section{Introduction}
\label{sec:introduction}

Phrase-based statistical machine translation (PBSMT) \cite{williams2016} has proven to be an effective approach to the task of machine translation. Even though in the last years neural systems have gained most of the attention from industry and academia, a number of recent works show that statistical approaches may still provide relevant results in hybrid, unsupervised or low-resource scenarios~\cite{artetxe,byrne}. In PBSMT systems, the source-language (SL) sentence is split into non overlapping word sequences (known as \emph{phrases}) and a translation into target-language (TL) is chosen for each phrase from a \emph{phrase table} of bilingual phrase pairs extracted from a parallel corpus. A decoder efficiently chooses the splitting points and the corresponding TL equivalents by using the information provided by a set of features, which usually include probabilities provided by translation models as well as a language model. In spite of their performance, PBSMT systems are affected by some limitations due to the local strategy they follow. More specifically, they tend to overpass long range dependencies, which may negatively affect translation quality. Additionally, phrase reordering is usually instrumented by means of distortion heuristics and lexicalised reordering models or an operation-sequence model \cite{durrani15_osm} that cannot cope with the multiple structural divergences that are often necessary to translate between most language pairs.

Tree-based models address these issues by relying on a recursive representation of sentences that allows for \emph{gaps} between words. Among these models, hierarchical phrase-based statistical machine translation (HSMT)~\cite{chiang2005,Chiang-CL-2007} has gained lot of attention due to its relative simplicity and the lack of need for linguistic knowledge. HSMT infers  synchronous context-free grammars (SCFG); remarkably, HSMT infers grammars with a single non-terminal $X$.\footnote{Strictly speaking, two non-terminals are used since an additional non-terminal (used in the glue rules) is set as the initial symbol of the grammar~\cite{chiang2005,Chiang-CL-2007}.} \newcite{Chiang-CL-2007} also sets some additional restrictions on the extracted rules in order to contain the combinatorial explosion in the number of rules, thus reducing decoding complexity.

This paper shows that the main limitation introduced in standard HSMT, namely, the use of a single non-terminal, can be overcome to improve translation quality. Our strategy differs from standard HSMT in that a different non-terminal is used for every possible gap when extracting the initial set of rules; this may easily result in millions of non-terminals which are then merged following an equivalence criterion, thus reducing the initial number of non-terminals in several orders of magnitude. Our experiments show that the resulting set of non-terminals correctly capture the contextual information that makes it possible to statistically significantly improve the BLEU score of the standard HSMT approach. However, additional research has to be carried out so that our method scales up with large training corpora.

Section~\ref{sec:sota} discusses related works. Section~\ref{sec:grammars} then introduces the formalism of synchronous context-free grammars and the standard procedure to obtain them from parallel corpora and use them in HSMT. Section~\ref{sec:method} presents our proposal for rule inference, including our criterion for variable equivalence and refinement. The experimental set-up and the results of our experiments are presented in Section~\ref{sec:experiments}. Finally, the paper ends with some discussion and conclusions.

\section{Related work}
\label{sec:sota}

Probabilistic context-free grammars (PCFGs) have been traditionally used to
model the generation of monolingual languages \cite{Charniak-1994}. The refinement of
probabilistic context-free grammars  has been addressed
before in a number of papers.  For example, \newcite{Matsuzaki-ACL-2005}
use a set of $N$ latent symbols to annotate ---and
therefore specialize--- non-terminals in a probabilistic context-free
grammar.  Every non-terminal $A$ in the
grammar is split into $N$ non-terminals and the probabilities of
the new productions are estimated to maximize the likelihood of the
training set of strings. 


In the approach for the refinement of PCFGs followed by
\newcite{Matsuzaki-ACL-2005}, the number of latent non-terminals must be
small ($N$ ranging from 1 to 16 in the experiments) since the training
time and memory grow very fast with $N$. The input trees are also binarized before the
estimation of the parameters to avoid an exponential growth in the
number of productions contributing to every sum of probabilities.
The method outperforms non-lexicalized approaches in terms
of parsing accuracy measured over the Wall Street Journal (WSJ) portion of the Penn treebank~\cite{Marcus-CL-1993,Marcus-HTL-1994}.

In contrast, \newcite{Petrov-ACL-2006} apply a more sophisticated
procedure for the specialization of the non-terminals in a PCFG\@. Instead
of generating all possible annotations, their hierarchical splitting
starts with the nearly one hundred tags in the WSJ corpus
(binarization of the parse trees is also applied here) and leads to
about one thousand symbols with a significant reduction in parsing
error rate. The procedure splits iteratively every symbol into two
sub-symbols and performs expectation-maximization optimization to estimate the probability
of the productions. 
This method allows for a deeper recursive partition
of some symbols (up to 6 times in the experiments with the WSJ corpus,
as reported by the authors).  Their results
showed that significant improvements in the accuracy of parsing can be
achieved with grammars which are more compact than those obtained
by~\newcite{Matsuzaki-ACL-2005}. Parsing with the enhanced grammar can
be accelerated with a coarse-to-fine scheme~\cite{Petrov-HLTC-2007}
which prunes the items analyzed in the chart according to the
estimations given by a simpler grammar where non-terminals are clustered
into a smaller number of classes.

While strings are usually not enough to identify a particular PCFG,
there exist methods which guaranteee convergence to the true grammar
when structural information is available. For example,
\newcite{Pereira-ACL-1992} apply expectation-maximization optimization to the identification of PCFGs from
partially bracketed samples (sentences with parenthesis marking the
constituent boundaries which the analysis should respect). The
probabilistic tree grammars obtained after the optimization are
non-deterministic (in the sense that multiple states are reachable as
the result of processing a subtree bottom-up) and their trees must be
binarized, a procedure that needs some linguistic guidance in order to
generate meaningful probabilistic models. 

The identification of probabilistic grammars has been addressed also
through state-merging techniques. For example, such techniques have
been applied to the case of regular grammars modelling string
languages by \newcite{Stolcke-NIPS-1992} and
\newcite{Carrasco-ITA-1999}. In these methods, the initial grammar has
one state or non-terminal per string-prefix in the sample and, then, the
procedure looks for the optimal partition of the set of non-terminals, an approach that is similar to the one we also follow in this work.
Convergence is guaranteed by keeping a frontier of states with all
strings whose prefixes have been already examined.

Experimental work on the identification of regular string languages on
sparse data~\cite{Lang-ICGI-1998} has shown the importance of
exploring earlier the nodes about which there is more information
available (a procedure known as \emph{evidence-driven}
merging). This result suggests that instead of using the
straightforward breadth-first search (the simplest which guarantees
convergence) one should consider more elaborate orders where those
pairs of subtrees in the frontier with a higher number of observations
are considered earlier.


Generalizing the merging techniques for string languages,
\newcite{Carrasco-ML-2001} define a procedure which identifies any
regular tree language by comparing pairs of non-terminals: each initial
non-terminal corresponds to a subtree in the sample and they are compared
in a breadth-first mode and merged when they are found to be
compatible.

As already commented in the introduction, \newcite{chiang2005} extended the phrase-based approach in statistical machine translation to tree-based translation by adapting the rule extraction algorithm to learn phrase pairs with gaps~\cite{Koehn2010}, and also presented a decoding method based on chart parsing, thus obtaining a working hierarchical phrase-based statistical machine translation system. \newcite{Chiang-CL-2007} then refined the method by introducing cube pruning to improve the efficiency of the chart decoder. Other authors~\cite{MaillettedeBuyWenniger2015,vilar2010,mylonakis2011} have introduced additional improvements to the rule learning procedure in the original proposal.

\section{Synchronous context-free grammars in machine translation}
\label{sec:grammars}

\emph{Transduction grammars}~\cite{Lewis-ACM-1968} have been used in SMT to model the hierarchical nature of translation which is implicit in the alignments between words in a bitext~\cite{Wu-CL-1997,Chiang-CL-2007}.  A \emph{bitext} or \emph{bilingual text} $B=(s_1,t_1)(s_2,t_2)\cdots (s_{|B|}, t_{|B|})$ consist of a finite sequence of sentence pairs, where every second component $t_i$, the \emph{target sentence}, is the translation of the first component $s_i$ in the pair, the \emph{source sentence}.

A \emph{word-level alignment} $A(s,t)$ annotates every sentence pair $(s,t)$ with a subset of $\{1,\ldots,|s|\}\times\{1,\ldots,|t|\}$ ---where $|s|$ and $|t|$ are the sentence lengths--- and provides those
pairs of word positions in the source and target sentence which can be linked together according to a translation model.

A \emph{transduction grammar} $G=(V, \mathcal{S}, \mathcal{T}, I, R, H)$ consists of a finite set of non-terminals $V=\{X_1,\ldots, X_N\}$, two sets of terminals ---here, $\mathcal{S}$ and $\mathcal{T}$ consist of segments of words contained in source and target sentences respectively---, an initial symbol $I\in V$, a finite set of production rules (\emph{rules} or \emph{productions}, for short) $R=\{r_1,\ldots,r_M\}\subset V\times (\mathcal{S}\cup V)^{+}\times(\mathcal{T}\cup V)^{+}$ and a set of $M$ one-to-one mappings $H=\{h_1,\ldots h_M\}$ with $h_m:\mathbb{N}\to \mathbb{N}$. For every rule $r_m=(X_n,\alpha,\beta)\in R$, $h_m$ couples every instance of a non-terminal in $\alpha$ and an instance of the same non-terminal in $\beta$ ---therefore, $\alpha$ and $\beta$ must have an identical number and type of non-terminals.  A production $r_m=(X_n,\alpha,\beta)$ in a transduction grammar $G$ will be written in the following as $X_n\to (\alpha,\beta)$ and their left and right
components as $X_n=\text{left}(r_m)$ and $(\alpha,\beta)=\text{right}(r_m)$, respectively.

The \emph{synchronous context-free grammars} introduced by~\newcite{Chiang-CL-2007} are transduction grammars whose productions have some restrictions:

\begin{enumerate}
    \item there is a single productive non-terminal $X$, in addition to the initial non-terminal $I$, that is, $V=\{I, X\}$;
    \item there is an upper limit on the length of the bilingual phrases of 10 words in either side;
    \item there is an upper limit of 2 in the number of non-terminals that can appear in a production;
    \item the size of the source side of production rules can have at most 5 terminals and non-terminals;
    \item the productions cannot contain contiguous non-terminals on the source side and they must include some lexical content (terminals); and 
    \item two \emph{glue rules} are defined to start derivations: $I\to (X,X)$ and $I\to (I X, I X)$.
\end{enumerate}

The transduction grammars employed by~\newcite{Chiang-CL-2007} restrict terminals to be in the set $\Phi(A, B)$ of bilingual phrase pairs obtained with the same extraction algorithm used in phrase-based SMT~\cite[sect.~5.2.2]{Koehn2010}: a \emph{bilingual phrase pair} or \emph{biphrase} is a pair in $\mathcal{S}\times \mathcal{T}$ which is consistent with the word alignments $A$ provided by a statistical aligner for the bitext $B$.\footnote{Essentially, a pair $(u,v)$ is a bilingual phrase pair if: $u$ and $v$ are segments in the source and target sentence, respectively; no word in $u$ is aligned to a word not in $v$ and vice versa; and at least one word in $u$ is aligned to a word in $v$.} The procedure also applies the following extraction rules:
\begin{itemize}
\item For every phrase pair $(u,v)\in\Phi(A,B)$, add a production $X\to (u,v)$ to $R$.
\item For every production $r_m\in R$ such that $r_m=X\to (\alpha_1 u\alpha_2,\beta_1 v\beta_2)$ with $\alpha_i\in(\mathcal{S}+V)^{*}$,  $\beta_i\in(\mathcal{T}+V)^{*}$, and $(u,v)\in\Phi(A,B)$, add the production $r_{M+1}=X\to (\alpha_1 X\alpha_2,\beta_1 X \beta_2)$ to $R$ ---with $h_{M+1}$ extending $h_m$ with a new link between the inserted $X$ pair.
\end{itemize}

The previous procedure ends up generating production rules such as the following English--Chinese rule:
$$
X\to (\mathrm{hyu} \,X^{[1]}\, \mathrm{you} \,X^{[2]} \,\, | \,\, \mathrm{have} \,X^{[2]}\, \mathrm{with} \,X^{[1]})
$$
where the numbers in the superindexes are not used to represent different non-terminals but the coupling between non-terminals resulting from the corresponding one-to-one mapping $h_i$.

Each rule is given a probabilistic score. In order to find the most probable translation of an input sentence according to the grammar model, chart parsing is used at decoding time~\cite{Chiang-CL-2007}.

\section{A new method for grammar induction}
\label{sec:method}

Our strategy differs from the one by~\newcite{Chiang-CL-2007} already introduced in the previous section in that a different non-terminal is used for every possible gap when extracting the initial set of rules; this may easily result in millions of non-terminals which are then merged following an equivalence criterion, thus reducing the initial number of non-terminals in several order of magnitudes. Consequently, the following sections present the algorithm for extraction of production rules (Section~\ref{sec:extraction}), 
the criterion for considering two non-terminals as equivalent (Section~\ref{sec:equivalence}) and the merging methods which join non-terminals based on the equivalence criterion (Section~\ref{sec:merging}).

\subsection{Extraction of production rules}
\label{sec:extraction}

The extraction phase assigns a different non-terminal to every production as follows (compare with the strategy proposed by~\newcite{Chiang-CL-2007} and described in Section~\ref{sec:grammars}):
\begin{enumerate}
\item Start with the set of initial non-terminals $V=\{I\}$ ---$I$ being the
  initial symbol--- and empty set of production rules $R=\emptyset$.
\item For every phrase pair $(u,v)\in\Phi(A,B)$, add a new non-terminal
  $X_n$ to $V$ ---$n$ being the current size of $V$---, and the new production
  $X_n\to (u,v)$, to $R$. If $(u,v)$ is a sentence pair, then add also
  $I\to X_n$ to $R$.  In contrast to~\newcite{Chiang-CL-2007}, the length of the phrase-pairs used is not constrained. This results in a large number of production rules as well as in a large number of non-terminals but it is necessary to be able to reproduce each sentence pair in the training corpus. 

\item For every production $r_k\in R$ such that $r_k=X_i\to (\alpha_0
  \alpha_1\alpha_2,\beta_0\beta_1\beta_2)$ and there is a
  production $X_j\to (\alpha_1,\beta_1)\in R$, add the production
  $r_{m+1}=X_i\to (\alpha_0 X_j\alpha_2,\beta_0X_j \beta_2)$ to $R$ ---$m$
  being the size of $R$. Note that the subscript in the left-hand side is not changed, that
    is, $\text{left}(r_{m+1})=\text{left}(r_k)$. Note also that for every phrase pair $(u,v)\in\Phi(A,B)$ there is only one non-terminal $X_n$ that can be derived to obtain $(u,v)$, either by means of the immediate rule $X_n\to (u,v)$ or through a number of derivations starting with some other rule $r_k \in R$ having $\text{left}(r_k)=X_n$. 
\item Finally, the count of every non-terminal $X_i$ is computed as done by~\newcite{Chiang-CL-2007}, that is, $C(X_i)$ is the number of occurrences in the training corpus of the phrase pair $(u,v)\in\Phi(A,B)$ generated by $X_i$. In order to generate the count for the production rule $c(r_k)$, the count $C(X_i)$ is equally distributed among all the productions $r_k$ generating that phrase pair, that is, between  $X_i \rightarrow (u,v)$ and the other productions with $X_i$ in the left-hand side added in step 3.
\end{enumerate}
Figure~\ref{tab:monotonic_prods} shows the resulting production rules and their fractional counts after applying our extraction procedure to the German--English sentence pairs (``das neue Haus'', ``the new house'') and (``das House'', ``the house''), assuming a monotonic alignment in which the $i$-th word of one sentence is aligned with the $i$-th word of the other sentence. Following~\newcite{Chiang-CL-2007}, rules with no lexical content, such as $X_1\to$ ($X_2X_3$, $X_3X_2$), have not been considered.

%
\begin{table}
  \centering\small
  \begin{tabular}{lc}
    \hline
    Production & Count\\
    \hline
    $I\to (X_1,X_1)$ & \gape{1}\\
    $I\to (X_7,X_7)$ & \gape{1}\\
    \hline
    $X_1\to$ (das neue Haus, the new house) &
    \gape{$\frac{1}{6}$}\\
    $X_2\to$ (das neue, the new) &
    \gape{$\frac{1}{3}$}\\
    $X_3\to$ (neue Haus, new house) &
    \gape{$\frac{1}{3}$}\\
    $X_4\to$ (das, the) &
    \gape{$2$}\\
    $X_5\to$ (neue, new) &
    \gape{$1$}\\
    $X_6\to$ (Haus, house) &
    \gape{$2$}\\
    $X_7\to$ (das Haus, the house) &
    \gape{$\frac{1}{3}$}\\
    \hline
    $X_1\to$ ($X_2$ Haus, $X_2$ house) &
    \gape{$\frac{1}{6}$}\\
    $X_1\to$ (das $X_3$, the $X_3$) &
    \gape{$\frac{1}{6}$}\\
    $X_1\to$ ($X_4$ neue Haus, $X_4$ new house) &
    \gape{$\frac{1}{6}$}\\
    $X_1\to$ (das $X_5$  Haus, the $X_5$ house) &
    \gape{$\frac{1}{6}$}\\
    $X_1\to$ (das neue $X_6$, the new $X_6$) &
    \gape{$\frac{1}{6}$}\\
    $X_2\to$ ($X_4$ neue, $X_4$ new) &
    \gape{$\frac{1}{3}$}\\
    $X_2\to$ (das $X_5$, the $X_5$) &
    \gape{$\frac{1}{3}$}\\
    $X_3\to$ ($X_5$ Haus, $X_5$ house) &
    \gape{$\frac{1}{3}$}\\
    $X_3\to$ (neue $X_6$, new $X_6$) &
    \gape{$\frac{1}{3}$}\\
    $X_7\to$ ($X_4$ Haus, $X_4$ house) &
    \gape{$\frac{1}{3}$}\\
    $X_7\to$ (das $X_6$, the $X_6$) &
    \gape{$\frac{1}{3}$}\\
    \hline
  \end{tabular}
  \caption{Productions and fractional counts for the monotonic  alignment of the sentence pairs (``das neue Haus'', ``the new house'') and (``das Haus'', ``the house''). The first group of rules includes the \emph{glue} rules; the second group includes those rules added in step 2 of the algorithm in Section~\ref{sec:extraction}; the third group is made up of the rules added in step 3. The fractional production counts, computed as described in step 4, are shown in the second column.}
  \label{tab:monotonic_prods}
\end{table}
\subsection{Determining equivalent non-terminals}
\label{sec:equivalence}

As will be presented in Section~\ref{sec:merging}, our method will merge pairs of equivalent non-terminals. We will denote with $X_i\sim X_j$ the fact that $X_i$ and $X_j$ are equivalent and, thus, they must be merged.  Then, $X_i\sim X_j$ implies that for all pairs of productions $r_m$ and $r_n$ which, for some $\alpha$ and $\beta$ in $(V\cup T)^{+}$ and $X_k$ and $X_l$ in $V$, have the form
$$
r_m = X_k \to \alpha X_i \beta
$$
and 
$$
r_n = X_l \to \alpha X_j \beta
$$
the following equality is (approximately) satisfied
\begin{equation}
  \label{eq:equivalence}
  \frac{c(r_m)}{C(X_i)}\approx \frac{c(r_n)}{C(X_j)}  
\end{equation}
and, recursively, $X_k\sim X_l$. The approximate matching between the above quotients is probabilistic in nature and must be defined therefore in terms of a stochastic test, such as the \newcite{Hoeffding-JASA-1963} bound. When using this test for proportion comparison, two proportions $c_1/C_1$ and $c_2/C_2$ are not statistically different if:
\[
\left| \frac{c_1}{C_1} - \frac{c_2}{C_2} \right| < \sqrt{\dfrac{- \log \dfrac{\alpha}{2}}{2 \dfrac{C_1 C_2}{C_1 + C_2}}}
\]
where $\alpha$ is the confidence level.\footnote{Following~\newcite{sebban}, we use a Fisher exact test as a back-off test in the experiments when the number of observations is small.} Note that the possible use of $\alpha$ in the proportion test is twofold: on the one hand, we may set $\alpha$ to a fixed value in order to test whether two proportions are statistically different; on the other hand, we may compute the value of $\alpha$ for which the test changes from true to false and use this value as a continuous measure of non-terminal \emph{dissimilarity}: after isolating $\alpha$ in the Hoeffding test equation and removing terms which are constant across different evaluations, we can easily arrive to a function $D$ that provides the dissimilarity of two non-terminals in a particular comparison context based on their respective counts:

\begin{equation}
  \label{eq:dissimilarity}
D(C_1,C_2,c_1,c_2) = { \dfrac{C_1 C_2}{C_1 + C_2} \left( \frac{c_1}{C_1} - \frac{c_2}{C_2} \right)}^2
\end{equation}

The dissimilarity of two variables is therefore obtained as the \emph{maximum} value of $D$ for all the contexts representing the different tests performed upon the variables as described in the previous algorithm.



\subsubsection{Example of equivalence computation} 

In order to gain some insight on the meaning of equivalence between
non-terminals, let us now consider, for instance, the comparison between
$X_3$ and $X_6$ in the example in Table~\ref{tab:monotonic_prods}, which can be considered plausible candidates for
equivalence since they both generate a noun phrase pair. The
fractional number of occurrences for $X_3$ and $X_6$ are given by
$C(X_3)=\frac{1}{3}+\frac{1}{3}+\frac{1}{3}=1$ and
$C(X_6)=2$. Note that
$C(X_6)$ receives contributions from both sentence pairs.

Non-terminal $X_3$ is only in $X_1\to (\text{das}\ X_3,\ \text{the}\ X_3)$
---with weight $\frac{1}{6}$--- and therefore, equivalence implies
that a production with right-hand side (das $X_6$, the $X_6$) must be
in $R$ with a weight which is consistent with Equation~\eqref{eq:equivalence}.
Indeed, $X_7\to (\text{das}\ X_6,\ \text{the}\ X_6)$ is in $R$ with
weight $\frac{1}{3}$: if $X_3\sim X_6$ both production will appear
with a relative frequency which must be similar to the relative
frequency for $X_3$ and $X_6$.  In this case,
\[
\frac{c(X_7\to (\text{das}\ X_6,\ \text{the}\ X_6))}
{C(X_6)} = \frac{1/3}{2} = 
\frac{1}{6}
\]
while
\[
\frac{c(X_1\to (\text{das}\ X_3,\ \text{the}\ X_3))}{C(X_3)} = \frac{1/6}{1} = \frac{1}{6} 
\]
Furthermore, the left-hand sides, $X_1$ and $X_7$ must be also
equivalent.\footnote{Since they only appear in $I$-productions, this
  is trivially true.}

The quotients above reflect that the word pair (Haus, house) has been
observed in two different contexts: after the biword (das, the) and
also following (neue, new); in contrast, the phrase pair (neue Haus,
new house) has been only observed after (das, the). This asymmetry
will lead to different values in the relative frequencies. Of course,
one cannot expect to draw definitive conclusions from such a tiny
bitext ---clearly, a much larger sample will be needed to extract
reliable estimates--- but the example illustrates how the frequency
estimates provide hints to differentiate between equivalent non-terminal
pairs and those which, in contrast, should remain distinct.

Once productions with $X_3$ on the right-hand side have been checked,
those with $X_6$ should be checked although, by the symmetry of the
test, only those that have no correspondent $X_3$-produc\-tion. In our
example, the tests will be
\[
\frac{c(X_1\to (\text{das neue}\ X_6,\ \text{the new}\ X_6)}{C(X_6)} \approx 0
\]
and 
\[
\frac{c(X_3\to (\text{neue}\ X_6,\ \text{new}\ X_6)}{C(X_6)} \approx 0
\]
and, thus, $X_1\sim X_3$.

Note that, if done carefully, recursion is always finite because the
comparison between non-terminals is consistent with the depth of the
subtree associated to each non-terminal.

An efficient algorithm has to be carefully designed in order to avoid duplicate calls when the equivalence between $X_i$
and $X_j$ is tested (since equivalence is a symmetric and transitive
relation).


\subsection{Merging variables}
\label{sec:merging}

Given the equivalence criterion presented in the previous section, an algorithm for non-terminal merging can be run in order to group equivalent non-terminals and reduce the initial number. We have evaluated two different algorithms: an adaptation of the Blue-Fringe algorithm and a $k$-medoids clustering algorithm. 

\subsubsection{The Blue-Fringe algorithm}
\label{sec:blue}

The following description, based on that
by~\newcite{Lang-ICGI-1998} of the procedure proposed
by~\newcite{Juille-AAAI-1998}, adapts the Blue-Fringe algorithm for the
identification of regular string languages to the case of transduction
grammars.

The procedure splits the set of non-terminals $V$ into three subsets: red (the kernel $K$ of mutually non-equivalent non-terminals), blue (the frontier $F$ being explored) and white (the subset $W=V-K-F$ of non-terminals with pending classification). At every iteration a non-terminal is removed from the frontier $F$ and either it becomes a new member of $K$ or it is merged with an equivalent one in $K$ (and, thus, removed from $V$). As will be seen immediately, after every addition or merge, some non-terminals in $W$ can be moved to $F$.

In order to avoid a possible infinite recursion in equivalence tests, non-terminals in $F$ must not produce any non-terminal in $K$ through derivation. This condition can be guaranteed if a non-terminal $X_n$ can only enter $F$ if all the content on the right-hand side of productions with the form $X_n\to(\alpha,\beta)$ is either lexical or already in $K$: this implies that any production $r_m$ such that $X_n\in F$ is in $\text{right}(r_m)$ satisfies $\text{left}(r_m)\in W$. 

Initially, leaves (non-terminals producing only irreducible phrase pairs) are red; non-terminals with a production with only lexical content and red non-terminals are blue; all other non-terminals are white. Our method will merge pairs of equivalent non-terminals by comparing those with a higher number of observations first~\cite{Lang-ICGI-1998}. The policy described by~\newcite{Juille-AAAI-1998} performs the following actions while there are still blue non-terminals:
\begin{enumerate}
\item Evaluate all red--blue merges.
\item If there exists a blue non-terminal that cannot be merged with any red non-terminal (meaning that no equivalent non-terminal is found), promote one of the shallowest such blue non-terminals to red (ties are broken at random).
\item Otherwise (if no blue non-terminal can be promoted), perform the red--blue merge with highest score. This score here is be based on the fractional number of occurrences of the corresponding non-terminals.
\end{enumerate}

Then, some white non-terminals are moved to the frontier $F$ as stated before.



\subsubsection{$k$-medoids clustering}
\label{sec:medoids}

The $k$-medoids algorithm is a clustering algorithm similar to the well-known $k$-means, but with the particularity that the center of each cluster (the \emph{medoid}) is a point in the data, which is specially relevant in our case since the representative of each cluster must be an existing non-terminal in the grammar. Unlike the Blue-Fringe algorithm, which attains a different number of final non-terminals depending on the confidence level $\alpha$, the parameter set \emph{a priori} in this case is the number of final clusters (i.e. non-terminals) $k$. Note that when following this merge approach, the dissimilarity between non-terminals in Equation~\eqref{eq:dissimilarity} is used to compute the required distances.

\section{Experimental setup}
\label{sec:experiments}

We trained and evaluated our grammar induction procedure on the English--Spanish language pair using a small fraction of the EMEA corpus.\footnote{\url{http://opus.nlpl.eu/EMEA.php}} Table \ref{tb:corpus} provide additional information about the corpora used in the experiments.

In order to have a manageable initial set of non-terminals and productions and make the problem computationally affordable we limited the sentences to be included in the training corpus to a maximum of 20 words. In addition, instead of using the words themselves when defining the initial set of non-terminals we used word classes. In particular, we used 10 word classes obtained  by running mkcls\footnote{\url{http://www.statmt.org/moses/giza/mkcls.html}} \cite{och99} for 5 iterations. As a result, the initial set of  non-terminals contains 852,423 non-terminals and the amount of productions, not including those involving the initial non-terminal $I$, is 2,055,902. 

\begin{table}
\centering
\begin{tabular}{crr}
\hline
Corpus & \# Sentences & \# Words (en/es)\\
\hline
train & 73,372 & 519,763 / 556,453\\
dev & 2,000 & 22,410 / 25,219 \\
test & 3,000 & 33,281 / 37,492 \\
\hline
\end{tabular}
\caption{Number of sentences and words in each language for the corpora used in the experiments.}
\label{tb:corpus}
\end{table}

All the experiments were carried out with the free/open-source SMT system Moses~\cite{koehn2007}, release 2.1.1. GIZA++~\cite{ochney2003} was used for computing words alignments.  KenLM was used to train a 5-gram language model on a monolingual corpus made of Europarl v7,\footnote{\url{http://www.statmt.org/wmt13/} \url{training-monolingual-europarl-v7.tgz}} News Commentary v8\footnote{\url{http://www.statmt.org/wmt13/} \url{training-monolingual-nc-v8.tgz}} and the EMEA sentences in the training corpus; in total the corpus used for training the language model consists of 3,423,702 Spanish sentences. The weights of the different feature functions were optimised by means of minimum error rate training~\cite{och2003}. The parallel corpora were tokenised and truecased before training, as were the development and test sets used.

\section{Results}

Table \ref{tb:bluefringe} reports the BLEU scores obtained on the test set when running the Blue-Fringe algorithm for different values of $\alpha$.  The amount of output non-terminals in the inferred context-free grammar is also reported. The best results are obtained with $\alpha=10^{-2}$. The performance of the baseline hierarchical phrase-based system \cite{chiang2005} is 0.5818. The difference in performance is statistically significant for the figures in bold according to paired bootstrap resampling \cite{koehn04:_statis} with $p=0.05$.  

\begin{table}
\centering
\begin{tabular}{cccc}
\hline
 $\alpha$ & \# final non-terminals  & BLEU \\
\hline
         $10^{-1}$ & 4,434 & 0.5838\\
         $10^{-2}$ & 2,346 & \textbf{0.5868}\\
         $10^{-3}$ & 1,606 & 0.5859\\
         $10^{-4}$ & 1,261 & 0.5855\\
         $10^{-5}$ & 1,074 & \textbf{0.5866}\\
\hline
\end{tabular}
\caption{BLEU scores obtained by the Blue-Fringe clustering algorithm for different values of $\alpha$. The performance of the baseline HSMT system is 0.5818.}
\label{tb:bluefringe}
\end{table}

Table \ref{tb:k-medoids} shows the BLEU scores obtained on the test set when the $k$-medoids clustering algorithm is run over the $n$ most frequent non-terminals (\#~clustered non-terminals) to obtained a pre-fixed number of clusters, that is, of non-terminals in the inferred grammar; the remaining non-terminals are then added to the nearest cluster after the algorithm finishes. The table reports results for 2, 3 and 4 clusters; although we tried with more clusters these are the numbers of clusters for which we got the best results.  The difference in performance with the baseline is statistically significant for all the figures reported in the table according to paired bootstrap resampling \cite{koehn04:_statis} with $p=0.05$.  

The results obtained when the number of non-terminals over which the k-medoids is run is set to 250 and the number of cluster to obtained is set to 3 are better than those obtained with the Blue-Fringe algorithm and better that the results achieved by the baseline. The k-medoids allow us to get an 3\% improvement in BLEU with just three non-terminals, in contrast  with the thousand non-terminals obtained by the Blue-Fringe algorithm.

\begin{table}
\centering
\begin{tabular}{cccc}
\hline
\multicolumn{1}{c}{\# clustered} & \multicolumn{3}{c}{\# final non-terminals $k$} \\
   non-terminals & 2 & 3 & 4 \\
\hline
125	 & 0.5975 & 0.5991 & 0.5952\\
250	 & 0.5986 & 0.6000 & 0.5942\\
500	 & 0.5946 & 0.5972 & 0.5936\\
1000 & 0.5950 & 0.5998 & 0.5939\\
2500 & 0.5985 & 0.5984 & 0.5964\\
5000 & 0.5947 & 0.5991 & 0.5947\\      
\hline
\end{tabular}
\caption{BLEU scores obtained by the $k$-medoids clustering method for different sizes of the subset of non-terminals over which the clustering is performed and for different numbers of clusters (i.e. final non-terminals). The performance of the baseline HSMT system is 0.5818.}
\label{tb:k-medoids}
\end{table}

\section{Conclusions}

This work extends the well-known algorithm for rule extraction in hierarchical statistical machine translation originally proposed by~\newcite{Chiang-CL-2007}. Our proposal allows for more than one non-terminal in the resulting synchronous context-free grammar, thus incorporating a specialisation in the resulting non-terminals. Our method works by initially creating a different non-terminal for every possible gap when extracting the initial set of rules; this may easily result in millions of non-terminals which are then merged following a novel equivalence criterion for non-terminals. Two merging strategies are proposed: one inspired on the Blue-Fringe algorithm that joins non-terminals on a one-by-one basis, and another one that performs a $k$-medoids clustering over a reduce set of non-terminals. A statistically significant improvement in BLEU as with respect to the original method is obtained with both merging criteria. 

For the experiments we used a small parallel corpus and restricted the length of the parallel sentences in the training corpus to 20 words. This was necessary in order to be able to run the merging algorithm in reasonable time. Recall that, in contrast to \shortcite{Chiang-CL-2007}, we do not restrict the length of the phrase pairs in order to be able to reproduce the parallel sentences in the training corpus; otherwise long-range reorderings happening near the root of the parse tree of the sentences would not be possible. We tried different methods for filtering the rule table before applying the approach described in this paper so as to be able to use larger corpora and longer sentences, although with no success. A deeper exploration of potential optimizations is necessary.

\paragraph{Acknowledgements}
Work supported by the Spanish government through project EFFORTUNE (TIN2015-69632-R). The authors thank Mikel L. Forcada for his helpful comments.


\bibliography{multihsmt}

\begin{thebibliography}{}

\bibitem[\protect\citename{Artetxe \bgroup et al.\egroup }2018]{artetxe}
Artetxe, Mikel, Gorka Labaka, and Eneko Agirre.
\newblock 2018.
\newblock Unsupervised statistical machine translation.
\newblock In {\em Proceedings of the 2018 Conference on Empirical Methods in
  Natural Language Processing}.

\bibitem[\protect\citename{Carrasco and Oncina}1999]{Carrasco-ITA-1999}
Carrasco, Rafael~C. and Jos{\'e} Oncina.
\newblock 1999.
\newblock Learning deterministic regular grammars from stochastic samples in
  polynomial time.
\newblock {\em Informatique Théorique et Applications}, 33(1):1--20.

\bibitem[\protect\citename{Carrasco \bgroup et al.\egroup
  }2001]{Carrasco-ML-2001}
Carrasco, Rafael~C., Jos{\'e} Oncina, and Jorge Calera-Rubio.
\newblock 2001.
\newblock Stochastic inference of regular tree languages.
\newblock {\em Machine Learning}, 44(1/2):185--197.

\bibitem[\protect\citename{Charniak}1994]{Charniak-1994}
Charniak, Eugene.
\newblock 1994.
\newblock {\em Statistical Language Learning}.
\newblock MIT Press.

\bibitem[\protect\citename{Chiang}2005]{chiang2005}
Chiang, David.
\newblock 2005.
\newblock A hierarchical phrase-based model for statistical machine
  translation.
\newblock In {\em Proc. ACL}, pages 263--270.

\bibitem[\protect\citename{Chiang}2007]{Chiang-CL-2007}
Chiang, David.
\newblock 2007.
\newblock Hierarchical phrase-based translation.
\newblock {\em Computational Linguistics}, 33(2):201--228.

\bibitem[\protect\citename{Durrani \bgroup et al.\egroup }2015]{durrani15_osm}
Durrani, Nadir, Helmut Schmid, Alexander Fraser, Philipp Koehn, and Hinrich
  Schutze.
\newblock 2015.
\newblock The operation sequence model -- combining n-gram-based and
  phrase-based statistical machine translation.
\newblock {\em Computational Linguistics}, 41(2):157--186.

\bibitem[\protect\citename{Habrard \bgroup et al.\egroup }2003]{sebban}
Habrard, Amaury, Marc Bernard, and Marc Sebban.
\newblock 2003.
\newblock Improvement of the state merging rule on noisy data in probabilistic
  grammatical inference.
\newblock In {\em Proceedings of {ECML} 2003, 14th European Conference on
  Machine Learning}, pages 169--180.

\bibitem[\protect\citename{Hoeffding}1963]{Hoeffding-JASA-1963}
Hoeffding, Wassily.
\newblock 1963.
\newblock Probability inequalities for sums of bounded random variables.
\newblock {\em Journal of the American Statistical Association},
  58(301):13--30.

\bibitem[\protect\citename{Juill{\'e} and Pollack}1998]{Juille-AAAI-1998}
Juill{\'e}, Hugues and Jordan~B. Pollack.
\newblock 1998.
\newblock A sampling-based heuristic for tree search applied to grammar
  induction.
\newblock In Mostow, Jack and Chuck Rich, editors, {\em Proceedings of the
  Fifteenth National Conference on Artificial Intelligence and Tenth Innovative
  Applications of Artificial Intelligence Conference, AAAI 98, IAAI 98}, pages
  776--783. AAAI Press / The MIT Press.

\bibitem[\protect\citename{Koehn \bgroup et al.\egroup }2007]{koehn2007}
Koehn, Philipp, Hieu Hoang, Alexandra Birch, Chris Callison-Burch, Marcello
  Federico, Nicola Bertoldi, Brooke Cowan, Wade Shen, Christine Moran, Richard
  Zens, Chris Dyer, Ond\v{r}ej Bojar, Alexandra Constantin, and Evan Herbst.
\newblock 2007.
\newblock Moses: Open source toolkit for statistical machine translation.
\newblock In {\em Proceedings of the 45th Annual Meeting of the ACL on
  Interactive Poster and Demonstration Sessions}, ACL '07, pages 177--180.
  Association for Computational Linguistics.

\bibitem[\protect\citename{Koehn}2004]{koehn04:_statis}
Koehn, P.
\newblock 2004.
\newblock Statistical significance tests for machine translation evaluation.
\newblock In {\em Proceedings of the 2004 Conference on Empirical Methods in
  Natural Language Processing}, pages 388--395.

\bibitem[\protect\citename{Koehn}2010]{Koehn2010}
Koehn, Philipp.
\newblock 2010.
\newblock {\em Statistical machine translation}.
\newblock Cambridge University Press.

\bibitem[\protect\citename{Lang \bgroup et al.\egroup }1998]{Lang-ICGI-1998}
Lang, Kevin~J., Barak~A. Pearlmutter, and Rodney~A. Price.
\newblock 1998.
\newblock Results of the {A}bbadingo {O}ne {DFA} learning competition and a new
  evidence-driven state merging algorithm.
\newblock In {\em Proceedings of the 4th International Colloquium on
  Grammatical Inference}, ICGI '98, pages 1--12. Springer-Verlag.

\bibitem[\protect\citename{Lewis~II and Stearns}1968]{Lewis-ACM-1968}
Lewis~II, P.~M. and R.~E. Stearns.
\newblock 1968.
\newblock Syntax-directed transduction.
\newblock {\em Journal of the ACM}, 15(3):465--488.

\bibitem[\protect\citename{Maillette~de Buy~Wenniger and
  Sima'an}2015]{MaillettedeBuyWenniger2015}
Maillette~de Buy~Wenniger, Gideon and Khalil Sima'an.
\newblock 2015.
\newblock Labeling hierarchical phrase-based models without linguistic
  resources.
\newblock {\em Machine Translation}, 29(3):225--265.

\bibitem[\protect\citename{Marcus \bgroup et al.\egroup }1993]{Marcus-CL-1993}
Marcus, Mitchell~P., Mary~Ann Marcinkiewicz, and Beatrice Santorini.
\newblock 1993.
\newblock Building a large annotated corpus of english: the {P}enn {T}reebank.
\newblock {\em Comput. Linguist.}, 19(2):313--330, June.

\bibitem[\protect\citename{Marcus \bgroup et al.\egroup }1994]{Marcus-HTL-1994}
Marcus, Mitchell~P., Grace Kim, Mary~Ann Marcinkiewicz, Robert MacIntyre, Ann
  Bies, Mark Ferguson, Karen Katz, and Britta Schasberger.
\newblock 1994.
\newblock The {P}enn {T}reebank: Annotating predicate argument structure.
\newblock In {\em Human Language Technology, Proceedings of a Workshop held at
  Plainsboro, New Jerey, USA, March 8-11, 1994}. Morgan Kaufmann.

\bibitem[\protect\citename{Matsuzaki \bgroup et al.\egroup
  }2005]{Matsuzaki-ACL-2005}
Matsuzaki, Takuya, Yusuke Miyao, and Jun'ichi Tsujii.
\newblock 2005.
\newblock Probabilistic cfg with latent annotations.
\newblock In Knight, Kevin, Hwee~Tou Ng, and Kemal Oflazer, editors, {\em ACL
  2005, 43rd Annual Meeting of the Association for Computational Linguistics,
  Proceedings of the Conference, 25-30 June 2005, University of Michigan, USA}.
  The Association for Computer Linguistics.

\bibitem[\protect\citename{Mylonakis and Sima{'}an}2011]{mylonakis2011}
Mylonakis, Markos and Khalil Sima{'}an.
\newblock 2011.
\newblock Learning hierarchical translation structure with linguistic
  annotations.
\newblock In {\em Proceedings of the 49th Annual Meeting of the Association for
  Computational Linguistics: Human Language Technologies}, pages 642--652.

\bibitem[\protect\citename{Och and Ney}2003]{ochney2003}
Och, Franz~Josef and Hermann Ney.
\newblock 2003.
\newblock A systematic comparison of various statistical alignment models.
\newblock {\em Comput. Linguist.}, 29(1):19--51.

\bibitem[\protect\citename{Och}1999]{och99}
Och, Franz~Josef.
\newblock 1999.
\newblock An efficient method for determining bilingual word classes.
\newblock In {\em Ninth Conf. of the Europ. Chapter of the Association for
  Computational Linguistics, EACL'99}, pages 71--76, June.

\bibitem[\protect\citename{Och}2003]{och2003}
Och, Franz~Josef.
\newblock 2003.
\newblock Minimum error rate training in statistical machine translation.
\newblock In {\em Proceedings of the 41st Annual Meeting on Association for
  Computational Linguistics}, pages 160--167.

\bibitem[\protect\citename{Pereira and Schabes}1992]{Pereira-ACL-1992}
Pereira, Fernando C.~N. and Yves Schabes.
\newblock 1992.
\newblock Inside-outside reestimation from partially bracketed corpora.
\newblock In Thompson, Henry~S., editor, {\em Proceedings 30th Annual Meeting
  of the Association for Computational Linguistics}, pages 128--135.

\bibitem[\protect\citename{Petrov and Klein}2007]{Petrov-HLTC-2007}
Petrov, Slav and Dan Klein.
\newblock 2007.
\newblock Improved inference for unlexicalized parsing.
\newblock In Sidner, Candace~L., Tanja Schultz, Matthew Stone, and ChengXiang
  Zhai, editors, {\em Proceedings Human Language Technology Conference of the
  North American Chapter of the Association of Computational Linguistics},
  pages 404--411. The Association for Computational Linguistics.

\bibitem[\protect\citename{Petrov \bgroup et al.\egroup }2006]{Petrov-ACL-2006}
Petrov, Slav, Leon Barrett, Romain Thibaux, and Dan Klein.
\newblock 2006.
\newblock Learning accurate, compact, and interpretable tree annotation.
\newblock In Calzolari, Nicoletta, Claire Cardie, and Pierre Isabelle, editors,
  {\em Proceedings ACL 2006, 21st International Conference on Computational
  Linguistics and 44th Annual Meeting of the Association for Computational
  Linguistics}. The Association for Computer Linguistics.

\bibitem[\protect\citename{Stahlberg \bgroup et al.\egroup }2016]{byrne}
Stahlberg, F., E.~Hasler, A~Waite, and B~Byrne.
\newblock 2016.
\newblock Syntactically guided neural machine translation.
\newblock In {\em Proceedings of the 54th Annual Meeting of the Association for
  Computational Linguistics}, pages 299--305.

\bibitem[\protect\citename{Stolcke and Omohundro}1992]{Stolcke-NIPS-1992}
Stolcke, Andreas and Stephen~M. Omohundro.
\newblock 1992.
\newblock {Hidden Markov Model} induction by bayesian model merging.
\newblock In Hanson, Stephen~Jose, Jack~D. Cowan, and C.~Lee Giles, editors,
  {\em Advances in Neural Information Processing Systems 5}, pages 11--18.
  Morgan Kaufmann.

\bibitem[\protect\citename{Vilar \bgroup et al.\egroup }2010]{vilar2010}
Vilar, David, Daniel Stein, Stephan Peitz, and Hermann Ney.
\newblock 2010.
\newblock If {I} only had a parser: Poor man's syntax for hierarchical machine
  translation.
\newblock In {\em Proc. of the Int. Workshop on Spoken Language Translation
  (IWSLT)}.

\bibitem[\protect\citename{Williams \bgroup et al.\egroup }2016]{williams2016}
Williams, Philip, Rico Sennrich, Matt Post, and Philipp Koehn.
\newblock 2016.
\newblock {\em Syntax-based statistical machine translation}.
\newblock Morgan \& Claypool Publishers.

\bibitem[\protect\citename{Wu}1997]{Wu-CL-1997}
Wu, Dekai.
\newblock 1997.
\newblock Stochastic inversion transduction grammars and bilingual parsing of
  parallel corpora.
\newblock {\em Computational Linguistics}, 23(3):377--403.

\end{thebibliography}
\bibliographystyle{multihsmt}

\end{document}